\documentclass[letterpaper, 10 pt, conference]{ieeeconf}

\IEEEoverridecommandlockouts                              
                                                          
\usepackage{graphics} 
\usepackage{epsfig} 
\usepackage{times} 
\usepackage{amsmath} 
\usepackage{amssymb}  
\usepackage[ruled,linesnumbered]{algorithm2e}
\usepackage{svg}
\usepackage{multirow}
\usepackage{array}
\usepackage{caption}
\usepackage{subcaption}
\usepackage{hyperref}
\usepackage{cite}

\usepackage[acronym]{glossaries}

\usepackage{acronym}

\usepackage{todonotes}

\title{\LARGE \bf
HIRO: Heuristics Informed Robot Online Path Planning \\ Using Pre-computed Deterministic Roadmaps
}

\author{Xi Huang$^1$, Gergely S\'{o}ti$^{1,2}$, Hongyi Zhou$^1$, Christoph Ledermann$^1$, Björn Hein$^{1,2}$, and Torsten Kröger$^1$
\thanks{The authors are with $^1$Institute for Anthropomatics and Robotics, Karlsruhe Institute of Technology, 76131 Karlsruhe, Germany, and 
$^2$Robotics and Autonomous Systems, Institute of Applied Research, Karlsruhe University of Applied Sciences, 76133 Karlsruhe, Germany\newline{\tt\small x.huang@kit.edu}}%
}

\begin{document}

\maketitle
\thispagestyle{empty}
\pagestyle{empty}

\begin{abstract}
With the goal of efficiently computing collision-free robot motion trajectories in dynamically changing environments, we present results of a novel method for Heuristics Informed Robot Online Path Planning (HIRO). %
Dividing robot environments into static and dynamic elements, we use the static part for initializing a deterministic roadmap, which provides a lower bound of the final path cost as informed heuristics for fast path-finding. These heuristics guide a search tree to explore the roadmap during runtime. The search tree examines the edges using a fuzzy collision checking concerning the dynamic environment. Finally, the heuristics tree exploits knowledge fed back from the fuzzy collision checking module and updates the lower bound for the path cost. As we demonstrate in real-world experiments, the closed-loop formed by these three components significantly accelerates the planning procedure. 
An additional backtracking step ensures the
feasibility of the resulting paths. Experiments in simulation and the real world show that HIRO can find collision-free paths considerably faster than baseline methods with and without prior knowledge of the environment.

\end{abstract}

\section{Introduction}
%
For a long time, research and development in robot motion planning were motivated by traditional industrial automation in structured environments. Sampling-based approaches are proven to be reliable and safe in these scenarios. For new types of applications, like human-robot collaboration, there are significantly different safety requirements.
As robots and humans share a workspace,
 the robots have to react to humans' movement instantly. Therefore, robots' ability to act in dynamic environments and answer frequently changing goal queries within a limited time is becoming more crucial.
For single-query methods, e.g., Rapidly Exploring Random Tree (RRT)\cite{lavalle1998rapidly}, which has a directional tree-like structure, changing the goal forces the algorithm to plan entirely from scratch. Multi-query methods, e.g., Probabilistic Roadmap (PRM)\cite{kavraki1996probabilistic}, on the other hand, compute a reusable roadmap and allow adding new start and goal configurations. 
After removing or deactivating the invalid edges, algorithms from both categories can find a new feasible solution on the modified roadmaps or trees. These approaches work fine for mobile or car-like platforms. However, applying these techniques to articulated robots is very expensive because there is no closed-form solution to map the changes in task space to a robot's configuration space. Therefore, sampling-based methods can usually not respond to environmental changes fast enough. As a result, it is common to slow down or pause robot motion until a new collision-free path for the changed environment is found. Recent approaches tried to address this problem with machine learning techniques, e.g. \cite{el2020towards}. However, these approaches usually simplify the problem to lower degree of freedoms and generally can not guarantee a safe trajectory.

 
\begin{figure}[!]
        \includegraphics[width=0.46\textwidth]{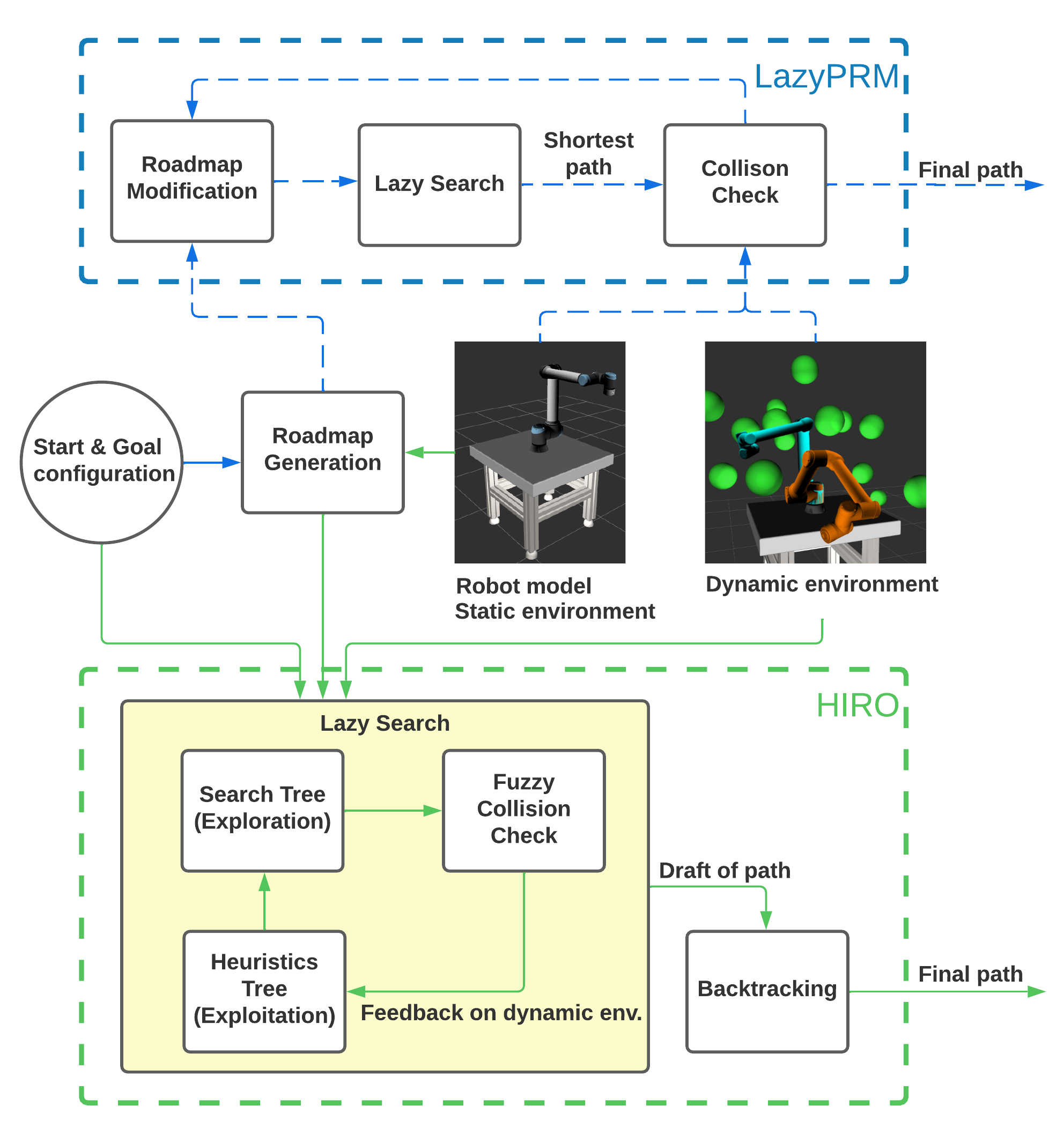}
    \caption{Our algorithm (HIRO, green) compared to LazyPRM\cite{bohlin2000path} (blue)}
    \label{fig:first_page}
\end{figure}

With the goal of quickly responding to the changes in the workspace, and keeping the advantages of sampling-based methods, e.g. collision-free motion,
we propose a novel path-planning method, called HIRO.
We divide the robot's workspace into a static and a dynamic environment. The static environment refers to permanently present objects, like a table where the robot is mounted or the robot itself. The dynamic environment contains everything else that alters, for example, manipulation objects or the user in a human-robot collaboration scenario. Similar to LazyPRM\cite{bohlin2000path}, our method generates a roadmap and then performs a lazy search for a solution. While LazyPRM ignores the robot's environment during its initial roadmap generation, HIRO uses a pre-computed roadmap 
to encode the static environment.

Within the lazy search, we propose the use of a closed-loop of three components, as shown in Fig. \ref{fig:first_page}, to speed up the process: A search tree selects those edges within the roadmap that has the best chance to be on the final path. These edges are examined via a fuzzy checking method based on a Jacobian-based safe zone concept. The so-called safe zones outline the regions where the planner is confident that are collision-free, given the information gained from the previous collision checks. This concept allocates the computational resources to the regions where we are not certain and therefore need checking. Then, a heuristics tree exploits the knowledge fed back from the fuzzy collision checking module and updates the lower bound for the path cost, which guides the search tree to focus on the edges that matter. To ensure that the resulting path is feasible, a backtracking step is employed to monitor and, if necessary, correct the final path.


Like LazyPRM and other multi-query planning methods, HIRO can respond to different planning requests without starting from scratch. Unlike LazyPRM, HIRO keeps the colliding nodes and edges in the roadmap by deactivating them only for the current search, which makes it possible to perform a lazy search using the same roadmap when the dynamic environment has changed. 
With HIRO, our major contributions are the following:
\begin{itemize}
    \item We use a pre-computed roadmap to distinguish the static and the dynamic environment. From this roadmap, we derive the informed heuristics which estimates the lower bound of the final path cost for the dynamic environment instead of the commonly used L2 distance.
    
    \item With a focus on dynamic changes, we form a closed loop to guide the exploration in the roadmap such that only relevant regions are searched. The closed-loop reduces the number of necessary edge evaluations by exploiting the knowledge obtained from the exploration and the collision checking.
    
    \item We develop a Jacobian-based collision checking method to accelerate the process of edge evaluation. It determines the safe zones and we can allocate the computational resources on the regions outside, where are not safe or we are uncertain about.

\end{itemize}

The rest of the paper is structured as follows:
Section \ref{sec::related_work} surveys related work. Details of the algorithm are described in Section \ref{sec:concept}. In Section \ref{sec:experiments}, the experimental setup and results are presented, where we show the efficacy of our method using numerous random scenarios as well as with a real-world setup. We address the future work in Section \ref{sec:discussion} and draw a conclusion in Section \ref{sec::conclusion}.

\section{Related Work}
\label{sec::related_work}

\subsection{Reactive Control}
The methods in this category generally try to solve collision avoidance problems on robot control level and provide real-time performance. A general procedure of these methods is to take the environmental changes as input and output a trajectory for the subsequent few time frames.
Artificial potential-fields\cite{khatib1986real} and numerous extended variations\cite{flacco2012depth} \cite{lee2021real} achieve real-time collision avoidance by exerting virtual forces on the robot to keep the robot away from obstacles. Further research like \cite{HaddadinUrbParBur2010} outlines a real-time method to handle unexpected virtual and physical contact. However, these methods can only serve as local planners and need a global planner to find their way to the goal.

\subsection{Roadmap Modification}
Sampling-based methods like RRT and PRM approximate the collision-free space with roadmaps and search for a valid path to answer the queries. 
Some parts of the node and edges on the roadmap become invalid on acquiring new information. In this case, modifying the roadmap may find a solution faster than planning from scratch.  
Dynamic $\text{A}^*$\cite{stentz1995focussed} marks the states affected by the changes in the environment and searches from the goal state to the robot's current state to avoid unnecessary updates of the roadmap. 
$\text{D}^*$-Lite\cite{koenig2002d} and its anytime version \cite{likhachev2005anytime} introduces the concept of lifelong planning and only updates the nodes that are relevant for computing the shortest path. Single query methods prune or rewire the tree to the root as invalid components are detected\cite{ferguson2006replanning} \cite{otte2016rrtx} \cite{zucker2007multipartite}. However, these methods are designed for and mainly examined with mobile platforms.  

For articulated robots, identifying the changes of task space in the configuration space is computationally expensive and generally needs long pre-processing. To enable the capability to modify or deactivate the roadmap in configuration space, DRM\cite{kallman2004motion} registers all samples in the configuration space to their belonging discrete cells in the task space. The corresponding samples and edges will be temporally deactivated if the environment alters. Distance-Aware Dynamically Weighted Roadmap (DA-DRM)\cite{knobloch2018distance} is a variant of DRM that considers the safety distance between robot and obstacles. Approaches like \cite{murray2016robot} encode the environment and the mapping on an FPGA chip to achieve fast planning.
Since these methods establish a look-up table between the configuration and task space, it takes up to hours to compute the mapping as preprocessing and some of these algorithms demand tremendous amount of memory. Different from building the look-up table between the task and configuration space, HIRO does not link the robot joint configurations to the occupied voxels in the task space and only keep a pre-computed roadmap in configuration space. This ensures us a light-weight pre-proccessing, which generally takes minutes instead of hours.

\subsection{Roadmap Generation}
Standard PRMs sample the nodes uniformly and connect them to their nearest neighbors, covering the collision-free space completely as the sample numbers increase. 
However, with the increasing number of nodes and edges in the roadmap, the information density of each unit decreases. The concept of roadmap sparsification\cite{coleman2015experience}\cite{dobson2014sparse} strives to generate a small-size roadmap that can fully represent the collision-free space applying to static environments.
Quasi-random roadmap\cite{branicky2001quasi} used Halton sequence to generate deterministic roadmaps, which have low dispersion and low discrepancy. These roadmaps appear to have regular neighborhood structures and offer a performance improvement compared to their randomized counterparts. A study on optimality and performance of deterministic roadmaps refers to \cite{janson2018deterministic}. The proposed approach is developed based on these features of pre-computed deterministic roadmaps.


\section{Informed Lazy Search in Pre-Computed Roadmap}
\label{sec:concept}
In a pre-computed roadmap, its edges and the nodes encode the collision-free space of the static environment. Given a start and goal query, we can find a feasible path without any collision checking regarding the static environment. The path cost in the static environment serves as a lower bound of the actual path cost in the dynamic environment, which is more realistic than the commonly used L2 distance between the current state and goal state.
We use these lower bounds as heuristics to guide us to efficiently traverse the roadmap and finally reach the goal. Furthermore, as we explore and gain more information from the current environment using the fuzzy collision check, we exploit the knowledge and update the heuristics cost. This helps us label the nodes with high expectations to be in the final solution path.

This section first exhibits the benefits of the pre-computed deterministic roadmaps and explains how to construct them. Then, we present the details of the three components in the lazy search, i.e., the search tree, fuzzy collision checking, as well as the heuristics tree as in Fig. \ref{fig:first_page}, and how do they interact with each other.


\subsection{Pre-Computed Roadmap}
\label{subsec:roadmap}
The basic idea of the pre-computed roadmaps is that they provide us with the benefit of knowing the actual path cost in the static environment, which can serve as the lower bound of the path cost considering dynamic environments.
Using pre-computed roadmaps to find solutions in dynamic environments, the roadmaps need to cover the collision-free space with low discrepancy. A lower discrepancy indicates that the samples cluster less and cover the space more homogeneously. Although adding more samples reduces the discrepancy, it results in a computational burden since more edges need to be evaluated. 
On the other hand, a deterministic roadmap has low discrepancy and regular connection structures, which facilitates approximating the free space with fewer samples and connections. For a comparison of the standard probabilistic roadmap and deterministic roadmap, please refer to \cite{branicky2001quasi} and \cite{janson2018deterministic}. Later in this work,  we use the deterministic roadmap generated via Halton sequences with two hyperparameters. They are the number of the connecting neighbours and the maximal length of a connection. Discussions on the hyperparameters and the optimization of roadmaps are out of the scope of this work.

\subsection{Heuristics for Roadmap Exploration}
 Useful heuristics is crucial for efficient search, especially if the edge evaluations are expensive. An example is the differences between A* and the Dijkstra algorithm.
We use a heuristics tree $\mathcal{T}_H$ to estimate the lower bound of the actual path cost from the given node to the goal. 
As the first step, we initialize $\mathcal{T}_H$ at the goal and run the classic A* algorithm towards the start. Collision checking does not take place in the entire search. 
Since A* is optimal, the connection in the heuristics tree between the start and the goal reflects the shortest possible route in the pre-computed roadmap. Upon unforeseen obstacles appearing in the workspace, the feasible path between the start and the goal will not be shorter. Hence, it can be seen as a lower path cost bound in this roadmap. This lower bound can be a much better estimation of the final path than merely using L2 distance.
Although collision checking is not involved in this step, running A* can introduce a large time consumption as the number of runs increases. Therefore, instead of repeating the A* algorithm from scratch, $\mathcal{T}_H$ grows incrementally to provide heuristics for the other nodes. 

We form the search guiding heuristics of a generic edge $h_e$ in the search tree $\mathcal{T}_S$ in two parts, the estimated number of edges to reach the goal $n_{reach}$ and the estimated path cost to reach $c_{reach}$. 
\begin{align}
h_e &= \{n_{reach}, c_{reach}\} \\
f_e &= \{n_{reach}, c_{reach} + c_{come}\}
\label{eq:f-score}
\end{align}
Like the A*, an f-score $f_e$ is pushed to the priority queue. While $h_e$ estimates the path cost in the future, $f_e$ is an expected cost of the whole path between the start and goal.
The priority queue sorts its items in lexicographic order. That is, it first compares $n_{reach}$, and then compares the $c_{reach} + c_{come}$ if the $n_{reach}$ of some items are identical. The details of lazy search using this heuristics is introduced in the following section.

\subsection{Lazy Search with Heuristics Updates}
In this section, we introduce the pipeline of the lazy search with the heuristics introduced above and how to update them as new knowledge of the environment is available. The concept of lazy search postpones the costly edge evaluation and only evaluates edges that matter. Evaluating edges refers to determining if an edge is collision-free. 

Our approach inherits the same philosophy of lazy searching. At a high level, given a start $v_s$ and goal $v_g$, we use two trees, $\mathcal{T}_S$ and $\mathcal{T}_H$ to find a path $\pi$ in the pre-computed roadmap $\mathcal{G}$.
In addition, we maintain a priority queue of edges $\mathcal Q_e$ to sort the f-score as in Eq. \ref{eq:f-score} in order. Edges with lower f-score are better and get examined sooner.

The search tree $\mathcal{T}_S$ growing from the start $v_s$ maintains the valid edges regarding the dynamic environment. It selects the edges from $\mathcal Q_e$ according to estimated path cost and examine them using the fuzzy collision checking module introduced in \ref{subsec:fuzzy}. Since all edges are collision-free for the static environment, the fuzzy collision checking module considers only the dynamic environments. The heuristics tree $\mathcal{T}_H$ grows from $v_g$, takes the information gained by the collision checking, and updates its estimation. The updated estimation helps $\mathcal{T}_S$ to focus on the relevant edges.
The closed-loop of these three components is the key to find a solution quickly. The complete pipeline of the search algorithm \ref{alg:search-pipeline} is described in the following.
We initialize both trees $\mathcal T_S$ and $\mathcal T_H$ and define a closed set $\mathcal{S}_{closed}$ (line 1-3). The closed set includes the nodes that have been confirmed to have a valid path from $v_s$, have a collision with the environment, or have no chance of being a part of the resulting path.
Then, we iterate the search loop (line 4-26) until a solution is found or the termination condition is met (line 4). The termination condition can be a limited number of iterations of the search loop or a specified search time. This is crucial for online applications. In the iteration, we first get the best edge based on the current knowledge of the environment (line 5). This step includes getting the best estimation from the priority queue $\mathcal{Q}_e$ and examining if this estimation is outdated. Since the estimations reflect the path cost in the static roadmap, they become outdated only if the collision checking has in between detected that at least one component on the path suggested by the heuristics tree is no longer valid. In this case, we update the heuristics as shown in Algorithm \ref{alg:heuristics-update}, continue the iteration from beginning(line 4) and consider the next best edge. 

Note that we keep a priority queue of the edges. The child node may have been already examined via other edges. If $v_c$ is in the closed set $\mathcal{S}_{closed}$, we skip this edge and enter the next iteration (line 7). Being in the closed set means that $v_c$ is either already in $\mathcal{T}_S$ or confirmed to be impossible to be on the final path.
Otherwise, we first examine if $v_c$ is valid regarding the dynamic environment (line 10). With $v_c$ being valid, we use the fuzzy collision checking method in section \ref{subsec:fuzzy} to verify the edge (line 15), which is the most expensive step of the algorithm. Upon the edge being valid, we expand the child node $v_c$ further (line 21) as described in Algorithm \ref{alg:search-expand}. Upon $v_c$ or the edge being invalid, we update the heuristics (line 12 and 24).

The update step described in Algorithm \ref{alg:heuristics-update} takes all the neighbours of the input node $v$ that are in $\mathcal{T}_H$ and selects the one with best estimated cost. If no neighbour is connected to $\mathcal{T}_H$, we grow $\mathcal{T}_H$ towards $v_c$ and finally update the heuristics. If no valid neighbours exist in the roadmap, this node will be added to the closed set and never be visited again. Updating the heuristics guides the search to avoid spending time on the edge evaluations that do not contribute to the result given the current knowledge of the environment. 
One of the key differences between our method and LazyPRM is that we update the heuristic instead of the roadmap when new environment information is obtained.

\begin{algorithm}
\SetAlgoLined
 $\mathcal{T}_S \gets v_s$, $\mathcal{T}_H \gets v_g$, $\mathcal{S}_{close}$ := $\{v_s\}$ \\
 Expand($v_s$, $\mathcal{G}$, $\mathcal{Q}_e$, $\mathcal{T}_S$, $\mathcal{T}_H$) \\
 $solution \gets False$ \\
 \While{$!solution \land !termination$}{
 $e \gets$ getBestEstimation($\mathcal{Q}_e$) \\
 $v_p \gets$ getParentNode(e), $v_c \gets$ getChildNode(e)  \\
 \If{$v_c \in \mathcal{S}_{close}$}{ Continue; }
 
 \If{\rm{!isValid}($v_c$)}
 {
    $\mathcal{S}_{close}$ := $\{\mathcal{S}_{close} , v_c\}$ \\
    \text{UpdateHeuristics}($v_c$, $\mathcal{G}$, $\mathcal{T}_H$) \\
    Continue;
 }
 \eIf{\rm{isValid}($e$)}
 {
    \If {\rm{inGoalRegion}($v_c$)}
    {
        $solution \gets True$; Break;\\
    }
    
    Expand($v_c$, $\mathcal{G}$, $\mathcal{Q}_e$, $\mathcal{T}_S$, $\mathcal{T}_H$) \\
    $v_c$.setPredecessor($v_p$, $\mathcal{T}_S$) \\
    $\mathcal{S}_{close}$ := \{$\mathcal{S}_{close}$ , $v_c$\} \\
 
 }{ \text{UpdateHeuristics}($v_c$, $\mathcal{G}$, $\mathcal{T}_H$) \\}
 }
\caption{Heuristics informed lazy search \label{alg:search-pipeline}}
\end{algorithm}
\begin{algorithm}
\SetAlgoLined
\SetKwInOut{Input}{Input}
\SetKwInOut{each}{each}
\Input{$v$, $\mathcal{G}$, $\mathcal{Q}_e$, $\mathcal{T}_S$, $\mathcal{T}_H$}
\For {$\text{\bf{each}}$ e in \rm{getOutEdges}($v$, $\mathcal{G}$)}
{
    $h_e$ := getHeuristics($e$, $\mathcal{T}_H$) \\
    $c_v$ := getCostToCome($v$, $\mathcal{T}_S$)  \\
    $f_e$ := getEstimatedCost($h_e$, $c_v$) \\
    $\mathcal{Q}_e$.push(e, $f_e$) \\
    $v_c$ := getChildNode(e) \\
}
\caption{Expand \label{alg:search-expand}}
\end{algorithm}
\begin{algorithm}
\SetAlgoLined
\SetKwInOut{Input}{Input}
\SetKwInOut{each}{each}
\Input{$v$, $\mathcal{G}$, $\mathcal{T}_H$}
\For {$\text{\bf{each}}$ $v_n$ $\rm {in}$ $\rm{getSuccessors}($v$, \mathcal{T}_H)$}
{
    \If{\rm{!hasNeighboursInTree}($v_n$, $\mathcal{G}$, $\mathcal{T}_H$)}
    {
       growHeuristicsTree($v_n$, $\mathcal{G}$, $\mathcal{T}_H$)
    }
    
    $v_{best} \gets$ getBestNeighboursInTree($v_n$, $\mathcal{G}$, $\mathcal{T}_H$) \\
        $v_n$.setPredessesor($v_{best}$, $\mathcal{T}_H$) \\
}
\caption{UpdateHeuristics \label{alg:heuristics-update}}
\end{algorithm}

\subsection{Fuzzy Collision Check}
\label{subsec:fuzzy}
The core idea of the fuzzy collision checking is to allocate the computational resources on the region where we are not certain.
Given a robot joint configuration, we determine its safe zone using the distance between the robot and the obstacles in the task space. Then, we skip the parts within the safe zones throughout the collision checking, where we are confident that they have a low chance of colliding, and check the uncertain parts outside the safe zones. For ensuring that the final path is feasible, we conduct a backtracking procedure. 

Determining the safe zone of a configuration starts with constraining the movement of the robot joints in certain directions. For the $i$-th obstacle in the environment, we can derive an equation for the closest point $C_{j,i}$ between this obstacle and the $j$-th link of the robot given its joint values $\mathbf{q} \in \mathbb R^{n}$, where $i, j \in \mathbb N$ and $n$ denotes the degrees of freedom of the robot. The relation between the torque $\tau \in \mathbb R^n$ in joint space and the Cartesian force $\mathbf{F}_{C_{j, i}} \in \mathbb R^3$ can be expressed with a partial Jacobian matrix $\mathbf{J}_{C_{j, i}}(q) \in \mathbb R^{3 \times n}$: 
\begin{equation}
\mathbf{\tau} = \mathbf{J}_{C_{j, i}}^T(\mathbf{q}) \mathbf{F}_{C_{j, i}}.
\end{equation}
 From now on, for the sake of clarity, we use $\mathbf{J}_{C_{j, i}}$ to represent $\mathbf{J}_{C_{j, i}}(\mathbf{q})$. By replacing the Cartesian force with the Cartesian distance vector $\mathbf d_{i,j}$, and the torque with a vector $\mathbf{s} \in \mathbb R^{n}$, we have
\begin{equation}
\mathbf{s} = \mathbf{J}_{C_{j, i}}^T \mathbf{d}_{j, i}.
\label{eq:s}
\end{equation}
The distance vector denotes the shortest distance from the $j$-th robot link to the $i$-th obstacle. The equation above implies the direction of the resulting joint torque and hence the joint movement if we pull the robot at $C_{j, i}$ along the distance vector $\mathbf{d}_{j, i}$. In other words, for the configuration $\mathbf{q}$, the sign of the vector $\mathbf{s}$ gives us the direction in which we should constrain to avoid a collision. So far, the constraints on the movement is similar to \cite{flacco2012depth}. Different from it, which constraints the velocity for every control cycle, we want to determine the free configuration spaces for the whole planning problem.
The next question is how far we should constrain the motion in every joint. Obviously, we can not handle the joints independently. In the following, we start with the instant Cartesian velocity of the configuration and derive an $n$-dimensional volume that represents the safe zone.
The instant velocity $\mathbf{v}_{C_{j, i}}$ of the point $C_{j, i}$ on the robot can be expressed with the joint velocity $\mathbf{\dot q} \in \mathbb R^n$ with a linearization at the joint configuration $\mathbf q$ as   
\begin{equation}
\mathbf{v}_{C_{j, i}} = \mathbf{J}_{C_{j, i}}\mathbf{\dot q}.
\end{equation}
With this linearization, we approximate the movement for a short time span $\Delta t$
\begin{equation}
\mathbf{v}_{C_{j, i}}\Delta t = \mathbf J_{C_{j, i}}\mathbf{\dot q}\Delta t. 
\end{equation}
By relating this movement to the distance vector $\mathbf d_{j, i}$, and using $\Delta \mathbf{q} = \mathbf{\dot q}\Delta t$, we form the necessary condition of a collision-free movement as following
\begin{equation}
||\mathbf{d}_{j, i}|| \geq ||\mathbf J_{C_{j, i}}\Delta \mathbf{q}||.
\label{eq:necessary_condition}
\end{equation}
The right-hand side of the inequality approximates the distance of the point $C_{j, i}$ if the joints have a displacement of $\Delta \mathbf{q}$.
Obviously, the relation above can not be used to determine the safe zone. In the following, we extend this solution to find a more conservative approximation of the collision-free space. The key to finding the conservative approximation is to identify the maximum Jacobian value along the robot link, which indicates the maximum velocity the robot can generate at this configuration $\mathbf{q}$. We assume that the links of the robot are straight, then the maximum Jacobian is located at one end of the link since the Jacobian changes linearly along the link.
Without losing the generality, we reform Eq. \ref{eq:necessary_condition} with a rotation matrix $R$ in $\mathbb R^3$
\begin{equation}
||R\mathbf d_{j, i} || \geq ||R\mathbf J_{C_{j, i}}\Delta \mathbf{q}||,
\label{eq:rotated_conditions}
\end{equation}
such that we can rotate the reference coordinate to align one of the bases, for example, $R\mathbf d_{j, i} = [||\mathbf d_{j, i}||, 0, 0 ]^T$. With this affine transformation, we simplify the constraint to 
\begin{equation}
    ||\mathbf d_{j, i}|| \geq \sum_{k=0}^j |J_{RC_{j, i}, k}| |\Delta q_k|,
\end{equation}
where $q_k$ denotes the $k$-th joint configuration and $|\cdot|$ is an absolute operator. $J_{RC_{j, i}, k}$ refers to the value of the $k$-th joint for the point $C_{j, i}$ in the matrix $R\mathbf J_{C_{j, i}}$. For the example above, these values correspond to the first row of $R\mathbf J_{C_{j, i}}$. 
Applying the maximum Jacobian to this constraints, we determine the safe zone by using simple linear programming
\begin{equation}
d_{j, i} \geq \sum_{k=0}^j |J_{RC_{j, i}, k, max}| |\Delta q_k|
\geq |\sum_{k=0}^j J_{RC_{j, i}, k} \Delta q_k|
\label{eq:linear_programming}
\end{equation}
where $d_{j,i} = ||\mathbf d_{j, i} || = ||R\mathbf d_{j, i} ||$ and $J_{RC_{j, i}, k, max}$ denotes the to maximum value regarding the $k$-th joint in the rotated Jacobian matrix along the $j$-th link.
\begin{figure}[!]
\centering
    \includegraphics[width=0.31\textwidth]{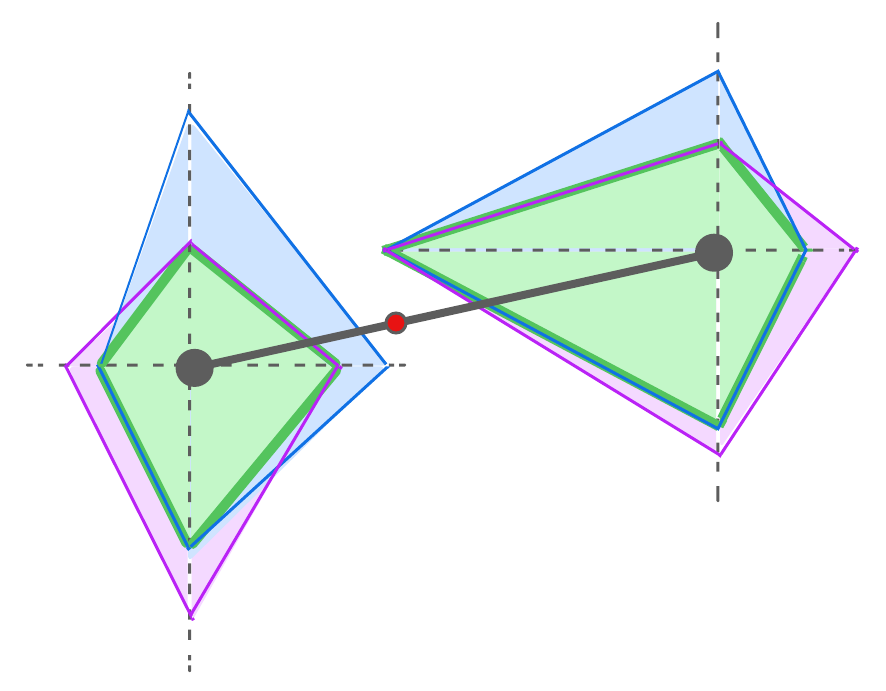}
    \caption{Safe zones (green) of two configurations (black dots) with two degrees of freedom; pink and blue quadrilaterals represent generic constraints determined by different links and obstacles; the green quadrilaterals, i.e., the safe zones, are formed by the strictest constraints from the pink and blue ones. The middle point of the part excluded by the safe zones is the next point on edge to check, marked as red. }
    \label{fig:safe_zone}
\end{figure}
The region determined by Eq. \ref{eq:linear_programming} is a close volume formed by hyperplanes. It restricts the motion of all joints symmetrically in both directions. Combining with Eq. \ref{eq:s}, we restrict the motion only in the direction that can cause collision. We specify the bound in the directions determined Eq. \ref{eq:s} regarding the $j$-th links and the $i$-th obstacle with 
\begin{align}
\Delta q_{{k, min}_{j, i}} &= -\frac{d_{j, i}}{|J_{RC_{j, i}, k, max}| + \epsilon}  \textrm{~ if ~} s_k < 0,\\
\Delta q_{{k, max}_{j, i}} &= \frac{d_{j, i}}{|J_{RC_{j, i}, k, max}| + \epsilon}  \textrm{~ if ~} s_k \geq 0,
\label{eq:single_constraint}    
\end{align}
where $\epsilon$ is a positive small value to handle the singularity. The bound in the opposite direction of Eq. \ref{eq:s} is set to be a constant, for example $\frac{\pi}{2}$. 
Applying the constraints to all links and obstacles, we use $\Delta q_{k, min}$ and $\Delta q_{k, max}$ to denote the approximated minimal and maximal allowed joint movement of the $k$-th joint, if other joints stay still. Using the linear programming in a similar form to Eq. \ref{eq:linear_programming}, we determine the hyperplanes with
\begin{align}
&\sum a_{j, max} \Delta q_{j, max} + b_{max} = 0, \\
&\sum a_{j, min} \Delta q_{j, min} + b_{min} = 0,
\end{align}
where $a_{j, max}$, $a_{j, min}$, $b_{max}$ and $b_{max}$ are coefficients for the hyperplanes. The hyperplanes form a close volume in the configuration space, which can be used as the safe zone.
An example in a configuration space with two degree of freedom is shown in Fig. \ref{fig:safe_zone}. In this example, the hyperplanes become lines of quadrilaterals. For a generic configuration $\mathbf q$, the test configuration $\mathbf q_{test}$ is in the safe zone, if 
\begin{align}
&\sum a_{j, max} (q_{test,j} - q_{j}) + b_{max} < 0, \\
&\sum a_{j, min} (q_{test,j} - q_{j}) + b_{min} < 0.
\end{align}
As shown in Fig. \ref{fig:safe_zone}, to verify the connection between two nodes, we first determine the safe zones of both nodes. The connection between two configuration in this case a straight line. Since we are certain that the part within the safe zone should be collision-free, we only need to focus on the part excluded by the both safe zones.
The checking process iterates by 
 computing safe zones of the middle points on the excluded segments, e.g. the red one in in Fig. \ref{fig:safe_zone}, until the edge is covered by safe zones or a collision is detected. 
 
To ensure the safety of the final path, we backtrack the edges on the final path. Once an edge on the final path is found to be invalid, we correct the solution. 
As the first step of the correction, we use the signed distances of the contact to recover the robot to a valid configuration. The outcome is a new configuration point, which bridges the two end configurations of the invalid edge. If no proper correction is found, we search for a new solution. Only a few new edges should be searched since the invalid edges and nodes are already deactivated. 


\section{Experiments}
\label{sec:experiments}
\begin{figure}
\centering
    \includegraphics[width=0.45\textwidth]{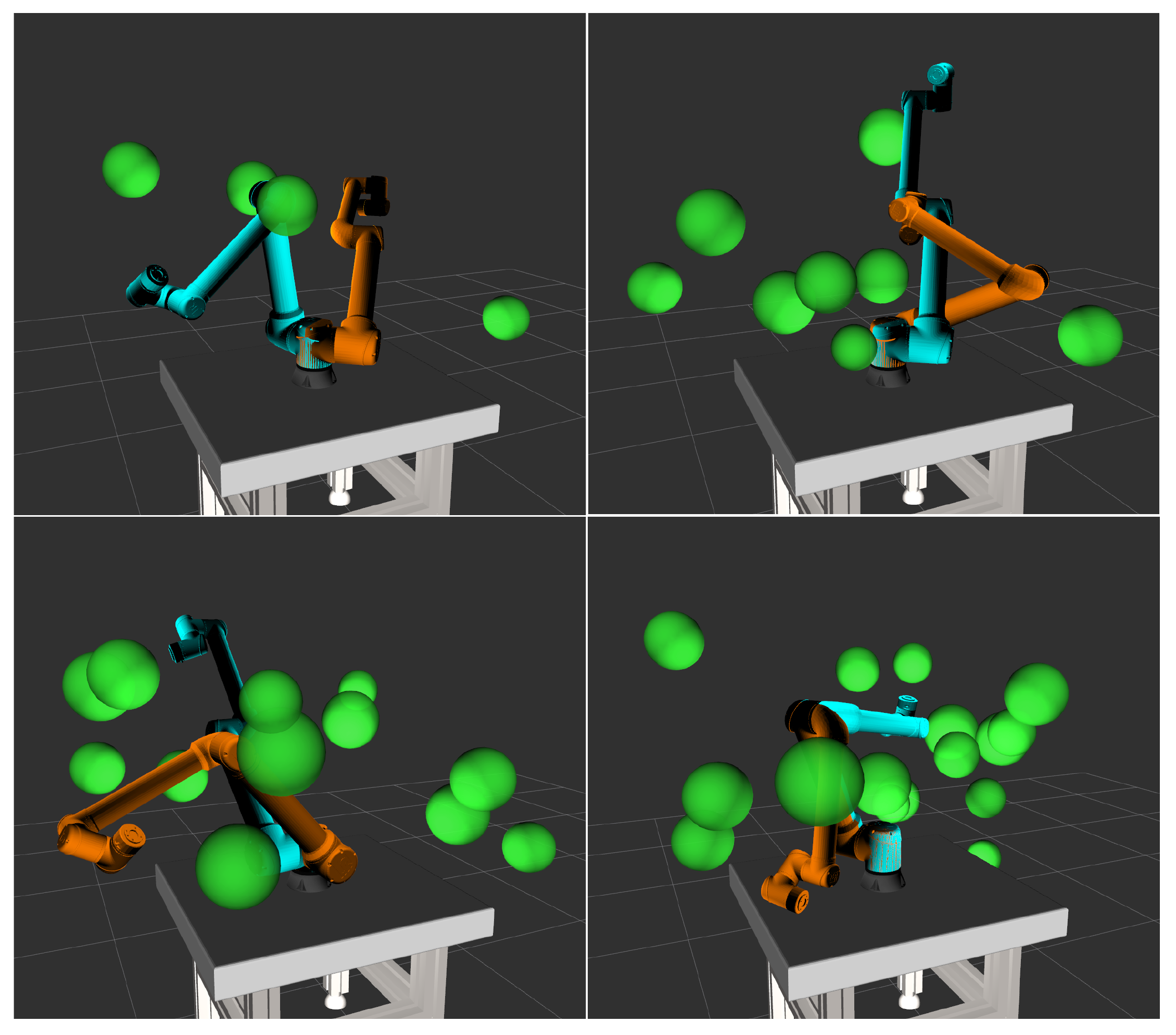}
    \caption{Examples of the planning scene datasets with 4, 8, 12 and 16 spherical random obstacles; start(orange) and goal(blue) poses are randomly generated.}
    \label{fig:planning_scenen_dataset}
\end{figure}

To examine the algorithm's efficacy and efficiency, we conduct both software and hardware experiments. For the software experiments, we generate datasets of random planning scenes . A planning scene includes unforeseen obstacles in the reachable workspace and a planning query with valid start and end configurations. At least one feasible path between the start and goal configuration exists in a planning scene. Fig. \ref{fig:planning_scenen_dataset} illustrates examples of the planning scene in the datasets. The green spheres are unforeseen objects for HIRO. The hardware experiment is conducted with an UR10e from Universal Robots. Throughout the hardware experiment, the robot has to respond to the environmental changes and adapt its motion accordingly, which is crucial for real-world applications in human-robot collaboration. The experiments are implemented with the Robot Operating System (ROS), Open Motion Planning Library (OMPL) \cite{sucan2012the-open-motion-planning-library}, and Flexible Collision Library (FCL) \cite{pan2012fcl}. In all three experiments, the results do not include path smoothing.

\subsubsection{Software experiment}
In the first experiment, we verify the improvement by comparing the planning results to the baseline methods. As baseline methods, we take open-sourced methods RRT, RRT-Connect, PRM, and LazyPRM from OMPL as in \cite{knobloch2018distance}. This experiment aims to simulate the scenarios in which both the goal and environments change rapidly, such that the planners have to plan a solution from the beginning. These scenarios usually happen as humans and robots share a workspace, or the sensor has a limited perception of the environment. We construct the deterministic roadmap once and use the same roadmap for all the planning scenes. The overhead of generating a roadmap will be introduced only once, depending on the size of the roadmap and the complexity of the static environment. In all three experiments, we use the roadmap with 40,000 nodes. Each node is allowed have 20 neighbours in a radius of $\frac{\pi}{2} \textrm{[rad]}$.

\begin{figure}[!]
    \centering
    \includegraphics[width=0.45\textwidth]{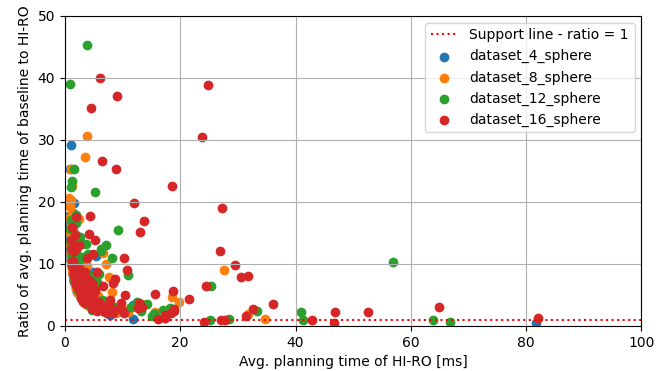}
    \caption{Improvements regarding the average planning time for different planning scenarios. RRTConnect is used as the baseline. Improvement above one implies that HIRO is better.}
    \label{fig:baseline_ratio}
\end{figure}

\begin{figure}[!]
\centering
    \includegraphics[width=0.45\textwidth]{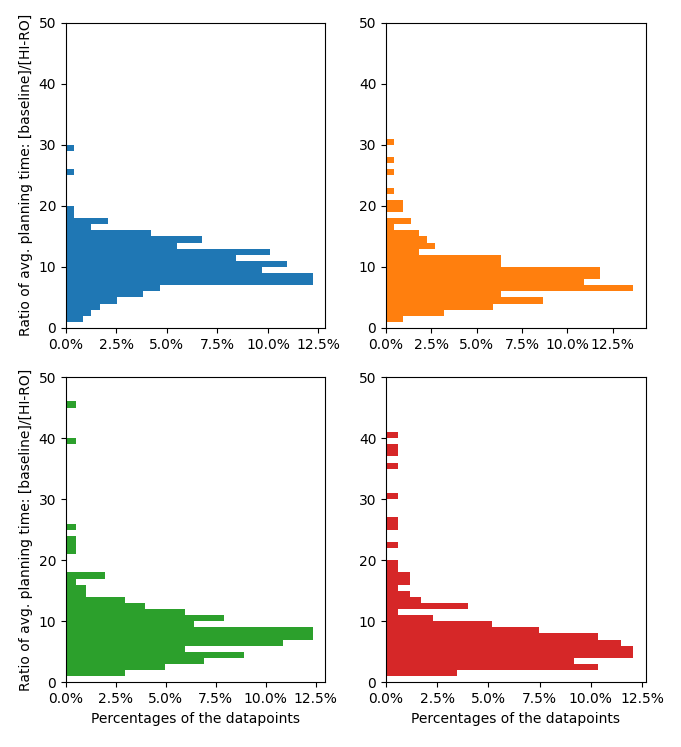}
    \caption{Histogram of the improvements regarding the average planning time for different planning scenarios. Dataset with 4, 8, 12 and 16 are shown in blue, orange, green and red.}
    \label{fig:baseline_ratio_hist}
\end{figure}

To represent the different complexity of the problems, we create four datasets of planning scenes with 4, 8, 12, and 16 obstacles. Every dataset includes 250 random scenes, where at least one feasible path exists between the start and end pose. 
Table \ref{table:results} lists the results of avg. planning time and standard deviation of the baseline methods and HIRO. 
In fact, the number of obstacles does not truly reflect the difficulty of a planning problem since difficulty depends on both the environment and the query. Fig. \ref{fig:baseline_ratio} presents the results of every planning scene as a dot. The vertical axis of the plot indicates how many times HIRO is better than the fastest baseline method RRT-Connect. For example, for the points that lay precisely on the red dashed support line, HIRO is identically fast as RRT-Connect in these planning scenes. The horizontal axis indicates the average planning time of HIRO. From this figure, we can easily see that HIRO plans a feasible path within 40 milliseconds for most of the scenes and is five to twenty times faster than RRT-Connect despite the outliers and extreme cases. The data points, where HIRO is more than 50 times faster than the baseline methods, are neglected for better and clearer data visualization.
The histograms in Fig. \ref{fig:baseline_ratio_hist} exhibit the distribution of the improvements. The vertical axes denote the ratio of the average planning time from the baseline method to HIRO. The data points whose ratio is over 50 are omitted, same as in Fig. \ref{fig:baseline_ratio}. With the number of unforeseen obstacles increasing, HIRO's improvement seems to decrease. Nevertheless, the dataset with 16 obstacles has a mean improvement of 7.62 times.


\begin{figure*}[h]
    \centering
    \includegraphics[width=0.95\textwidth]{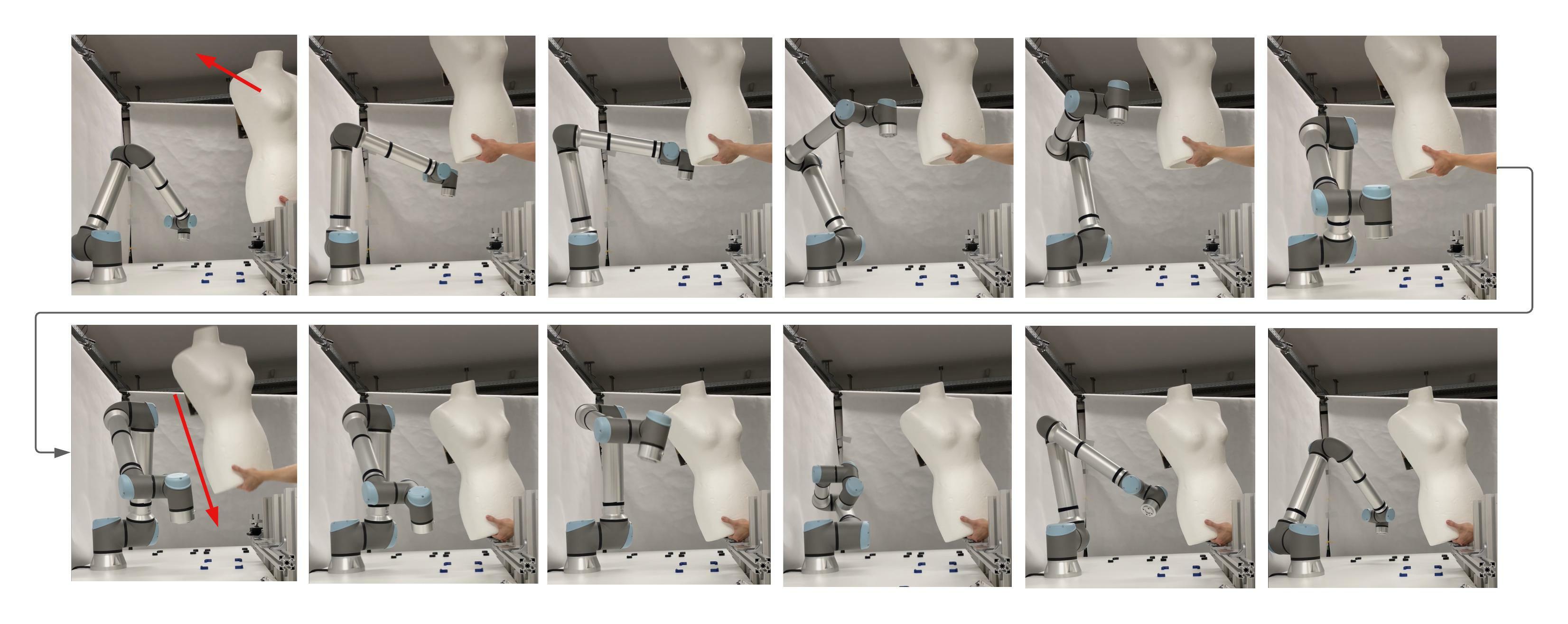}
    \caption{Hardware experiment with a human-size model Jessica moving within the workspace. The robot moves between the further end (top left) and the closer end of the table (bottom left). The red arrows indicate the sudden movement of Jessica while the robot is executing its motion. Snapshots are shown in chronological order from left to right.}
    \label{fig:robot_setup}
\end{figure*}
\begin{table*}[!]
\centering
\begin{tabular}{|l|c|c|c|c|c|c|c|c|}
\hline
\multicolumn{1}{|c|}{} & \multicolumn{2}{c|}{4 spheres} & \multicolumn{2}{c|}{8 spheres} & \multicolumn{2}{c|}{12 shperes} & \multicolumn{2}{c|}{16 spheres}\\
\cline{2-9}
\multicolumn{1}{|c|}{Methods} & mean & std & mean & std & mean & std & mean & std \\
\hline
RRT & 61.356 & 258.83 &83.249 &291.08 &147.90 &465.86 &230.09 &566.33 \\
RRTConnect &15.186 & 8.9953 &21.059 &26.459 &38.430 & 139.63 &58.360 &144.35\\
PRM & 46.526 & 188.41 &68.949 &185.46 &110.52 &320.10 & 222.25 &525.96 \\
LazyPRM & 27.828 & 209.81 & 33.731 & 104.28 & 66.974 & 239.57 & 177.71 & 838.46 \\
HIRO &2.4685 &6.4502 &5.3518 &17.974 &6.1051 &10.562 &11.959 &23.503\\
\hline
\end{tabular}
\caption{\label{table:results} Planning results of HIRO and baseline methods; mean and standard deviation are shown in milliseconds }
\end{table*}



\subsubsection{Hardware Experiment}
In the hardware experiment, we use a human-size model, Jessica, to represent the dynamic changes in the environment. Jessica appears and disappears in the workspace, making the robot have to adapt its motion frequently. Jessica is tracked by an ArUco \cite{garrido2016generation} marker via a camera.
Throughout the experiment, the robot traverses between both ends of the table. This scenario simulates that the robot shares the workspace with Jessica and primarily conducts a pick-and-place task. Fig. \ref{fig:robot_setup} presents a complete cycle of the workflow. The arrows refer to the sudden displacement of the model while the robot is moving. As a certain movement is detected, a planning request is triggered. To better visualize the planned path, we let the robot execute the adjusted path to the end. No path smoothing or shortening method is used for the presented path. The hardware experiment confirms that HIRO can quickly respond to environmental changes and adapt the path accordingly. A supplementary video of the hardware experiment is available at \url{https://youtu.be/ITBz1W7Ecbg}.



\section{Discussion}
\label{sec:discussion}
Software and hardware experiments show that HIRO can be a planner that reacts to environmental changes and plans globally. However, there are some limitations to be addressed in future works. The first limitation comes from the roadmap. We do not apply any filtering 
during the roadmap generation. Being aware of the clearance and Cartesian constraints and then filtering out nodes and edges with regard to certain rules can improve the quality of roadmaps. This can avoid sub-optimal trajectories and ease the burden of the follow-up trajectory generation algorithms.
The second limitation is that we do not use the knowledge from the previous planning. Learning form the experience to generate optimal roadmaps can mitigate the issue caused by limited roadmap resolution.
\section{Conclusions}
\label{sec::conclusion}
We proposed HIRO, an online path planning method designed to work in changing environments. HIRO uses a pre-computed collision-free roadmap with regard to the static environment as a basis for planning. This roadmap has to be computed only once and can be used for different planning queries. A closed loop, formed by a search tree, fuzzy collision checking and a heuristics tree, explores the free space and exploits the knowledge gain from the dynamic environment, which enables path adaptation to changes in the dynamic environment. Software and hardware experiments demonstrate that HIRO outperforms baseline methods and can be applied to read-world applications.

\section*{Acknowledgement}
\label{sec::acknowledgement}
The German Federal Ministry for Economic Affairs and Climate Action (BMWK) supported this work within the research project “FabOS” under grant 01MK20010B.




\bibliographystyle{IEEEtran}
\bibliography{bib}
\end{document}